\pdfoutput=1

\documentclass[11pt]{article}

\usepackage[]{acl}

\usepackage{times}
\usepackage{comment}
\usepackage{latexsym}
\usepackage{graphicx}
\usepackage[T1]{fontenc}

\usepackage[utf8]{inputenc}

\usepackage{microtype}

\usepackage{inconsolata}

\usepackage{booktabs}
\usepackage{caption}
\usepackage{subcaption}

\title{\textsc{CantonMT}: Cantonese to English NMT Platform with Fine-Tuned Models using Real and Synthetic Back-Translation Data}

\author{
           Kung Yin Hong, Lifeng Han, Riza Batista-Navarro, Goran Nenadic \\ \vspace*{0.075cm}
             Department of Computer Science, The University of Manchester\\ 
             Oxford Rd, Manchester M13 9PL, United Kingdom \\
            {\tt kenrick.kung@gmail.com }\\
            {\tt \{lifeng.han, riza.batista, g.nenadic\}@manchester.ac.uk} \\
}

\begin{document}
\maketitle
\begin{abstract}
Neural Machine Translation (NMT) for low-resource languages remains a challenge for many NLP researchers. In this work, we deploy a standard data augmentation methodology by back-translation to a new language translation direction, i.e., Cantonese-to-English. 
We present the models we fine-tuned using the limited amount of real data and the synthetic data we generated using back-translation by three models: OpusMT, NLLB, and mBART.
We carried out automatic evaluation using a range of different metrics including those that are lexical-based and embedding-based.
Furthermore, we create a user-friendly interface for the models we included in this project, \textsc{ CantonMT}, and make it available to facilitate Cantonese-to-English MT research. Researchers can add more models to this platform via our open-source\textsc{ CantonMT} toolkit, available at \url{https://github.com/kenrickkung/CantoneseTranslation}.



\end{abstract}

\section{Introduction}

Cantonese is one of the most popular dialects of Chinese languages, after the standard language Mandarin (the current official language in China, originally from the Beijing area), originally from the capital of Guangdong province, Guangzhou (a.k.a. Canton) in China. The population of Guangdong province was 129.51 million in 2022 according to the National Bureau of Statistics of China
\footnote{\url{https://data.stats.gov.cn/english}}. In addition, Cantonese is also the native language in Hong Kong (HK) and Macau regions which have populations of 7,503,100 and 704,149 in 2023,
according to HK Census and Statistics Department\footnote{\url{https://www.censtatd.gov.hk/en/}} and Macrotrends Global Population statistics.\footnote{
\url{https://www.macrotrends.net/global-metrics/countries/MAC/macao}} Furthermore, because of the economic growth in Guangdong, HK and Macau, many people from other Chinese provinces also learned to speak Cantonese for job purposes and due to cultural influences. There is also a large global population outside of China speaking Cantonese, 85.5 million, according to the Cantonese Language Association (CLA) \footnote{\url{https://cantoneselanguageassociation.byu.edu/}}.
In the era of the fast development of natural language processing (NLP), many machine translation (MT) models have been proposed for the majority of languages worldwide. However, \textit{low-resource language MT} remains a challenge for researchers. Cantonese translation using MT, specifically, is under-explored and has not been given much attention thus far.

In this work, we investigate one of the more popular MT methods, i.e. synthetic data augmentation via back-translation and model fine-tuning, as an approach to Cantonese-to-English neural MT (NMT), along the way introducing Cantonese-English as a new language pair.
We select several models for evaluation including both smaller and larger language models, and  compare their system performance using a range of evaluation metrics. 
Furthermore, we open-source our toolkit and create a web-based user-friendly platform called \textbf{CantonMT} to facilitate research on Cantonese-English translation. A public video demo is available.\footnote{\textsc{CantonMT} demo  \url{https://youtu.be/s8P5fJjS7Ls}}

In the next section (Section \ref{sec:RW}), we survey related work on Cantonese-English MT, data augmentation for MT, and available demos/engines. Section \ref{sec:exp} introduces our methodology and framework. Section \ref{sec:cantonMT} explains the web-based\textsc{ CantonMT} platform. Section \ref{sec:discuss} concludes this work with a discussion. 

\section{Related Work}
\label{sec:RW}

Research work focussing on Cantonese-English MT has not gained much attention to date. Earliest efforts include the work of \newcite{wu-etal-2006-structural} where example-based and rule-based MT were investigated. 
In recent years, a project plan on Cantonese-English Translation was put forward by researchers at the University of Hong Kong (HKU) where they proposed to investigate various MT approaches, including rule-based MT (RBMT), example-based MT (EBMT), statistical MT (SMT), gated-recurrent units (GRU) and transformers \cite{wing2020machine}.
More loosely related work include research in MT for Cantonese, but without English as the target language.
These include dialectal translation between Cantonese and Mandarin Chinese by \newcite{zhang-1998-dialect}, \newcite{Mak-etal-2012-LowResNMT-HK} and  \newcite{liu-2022-low}.

Data augmentation via backtranslation has been one of the standard practices for generating a synthetic corpus for improving MT performance on low-resource language pairs. This has been popular for both statistical MT (SMT) and NMT \cite{sugiyama-yoshinaga-2019-data,graça2019generalizing,edunov2020evaluation,nguyen2021cross,dataAugment2023NMT}. However, to the best of our knowledge, none of these efforts focused on Cantonese-to-English translation. 

Existing platforms or off-the-shelf demos for Cantonese-to-English MT are very scarce. Popular MT engines from commercial IT companies, including Google Translate\footnote{\url{https://translate.google.com}} and DeepL Translator,\footnote{\url{https://www.deepl.com/translator}} do not include this language pair. Both of them only included simplified and traditional characters of Mandarin Chinese.
Meanwhile, Microsoft Bing Translator\footnote{\url{https://www.bing.com/translator}} and Baidu, an IT company from China, made the Baidu Translator (Fanyi)\footnote{\url{https://fanyi.baidu.com/}} available, which includes Cantonese among several Chinese dialectal languages.\footnote{All these websites were last visited 4th March 2024.}
In the opposite direction, there are open-source tools for English-to-Cantonese MT from 
TransCan.\footnote{\url{https://github.com/ayaka14732/TransCan}}

\section{Experimental Work}
\label{sec:exp}
We introduce the methodology of CantonMT, experimental evaluations using the initial 38K 
real bilingual corpus, and extended model evaluations when we acquired 14.5K and 10K more real bilingual data from different sources subsequently. 

\begin{figure*}[]
\centering
\includegraphics[width=0.7\textwidth]{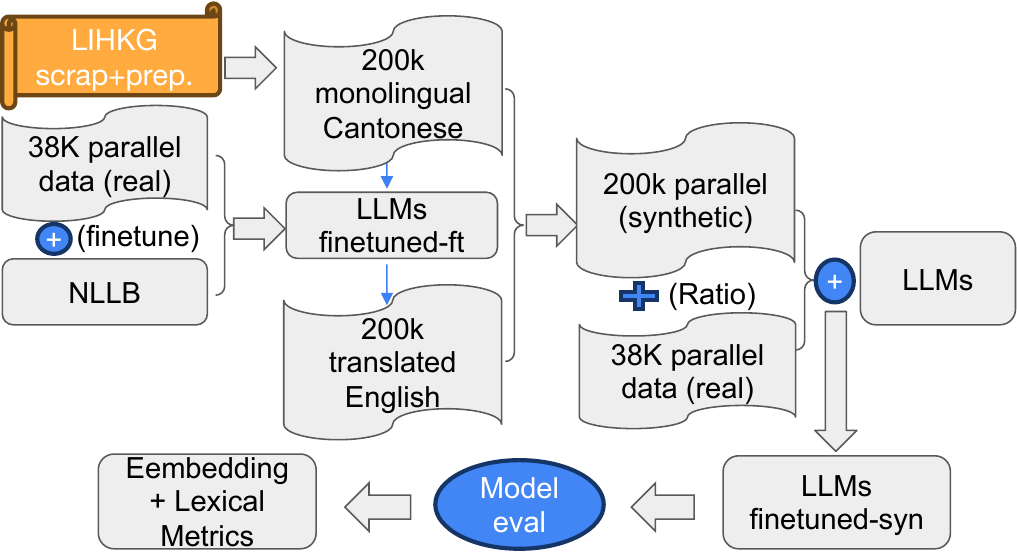}
    \caption{\textsc{CantonMT }Pipeline: data collection and preprocessing, synthetic data generation, model fine-tuning, model evaluation}
    \label{fig:CantonMT-pipeline-FineTune-cropped}
\end{figure*}

\subsection{Methodology and Framework}

The methodology of this work is presented in Figure \ref{fig:CantonMT-pipeline-FineTune-cropped}, which includes the following steps:
\begin{enumerate}
    \item DataPrep: data collection and pre-processing
    \item ModelFineTunePhase1: model selection for initial translator fine-tuning (ft, v1)
    \item SynDataGenerate: synthetic data generation using the initial translator and cleaned data
    \item ModelFineTunePhase2: second step MT fine-tuning using real and synthetic data (ft-syn)
    \item ModelEval: model evaluation using both embedding-based metrics (BERTscore and COMET) and lexical metrics (SacreBLEU and hLEPOR)
\end{enumerate}

For data collection, we scraped the data from the public Hong Kong forum LIHKG,\footnote{\url{https://lihkg.com}} which was launched in 2016 and has multiple categories including sports, entertainment, hot topic, gossip, current affairs, etc.
We extracted more than 1 million sentences from this website; however, the raw data comes with a lot of noise that needs to be cleaned, an example of which is shown in Figure \ref{fig:LIHKG-example} of Appendix \ref{sec:appendix}.
We carried out data cleaning to reduce noisy strings as well as data \textit{anonymisation} by removing user IDs from the text.
We also filtered out the sentences that were too short, i.e., with less than 10 Chinese characters. In the end, we prepared 200K clean monolingual Cantonese sentences for parallel synthetic data generation purposes. We shuffled the data for model training. 

In model fine-tuning phase 1, we aim to train a set of reasonable Cantonese-English MT models for synthetic data generation and model comparisons. 
The baseline models we selected are Opus-MT, NLLB and mBART. These were chosen to answer the following questions: (1) How much does model size impact fine-tuning performance? For this, we use Opus-MT which is a much smaller model trained on the Opus corpus using the MarianMT framework and NLLB-200, a very large language model pre-trained on 200+ languages from Mata-AI; (2) To what extent does it matter if the pre-trained translation models are exposed to Cantonese in their pre-training?
For this, we add mBART (\texttt{mbart-large-50-many-to-many-mmt}) which is another LLM but without Cantonese in its pre-training, vs NLLB which includes Cantonese.
Because the full-size NLLB is too large, we used the distilled model \texttt{nllb-200-distilled-600M}.



We fine-tuned these models using the available bilingual data from a bilingual Cantonese-English dictionary called ``Yue-Dian'',\footnote{\url{https://words.hk}} 
which is in total 44K in size. 
We divided this data into training, development and testing sets with 38K, 3K and 3K as their respective sizes, in light of the fact that the shared tasks organised by the Workshop on Statistical Machine Translation (WMT) tend to include around 3K sentences in their test sets.

In Step 3, synthetic data generation, we used the fine-tuned LLMs (LLM-ft-v1) from Step 2 to translate the monolingual Cantonese text we collected in Step 1. In this way, we obtain 200K back-translated English sentences; these synthetic sentences together with the Cantonese sentences create the 200K synthetic parallel corpus we generated. From now on, we will refer to the synthetic parallel corpus as \texttt{200K-ParaSyn}. 

In Step 4, we apply different ratios on the real parallel data we have at hand and on 200K-ParaSyn for LLM fine-tuning. We also test the influence of model switches, i.e. using different types of LLMs for LLM-ft (Phase 1) and LLM-syn (Phase 2).

In the last step, we deploy the fine-tuned LLMs in Phase 2 (LLM-syn) on the same test data and compare the results with LLM-ft (Phase 1) and {baseline models without fine-tuning.}
We also report comparisons with commerically available translation engines such as the Baidu Translator, Bing Translator and GPT4. 
The implementation of GPT-4 that we used is Cantonese Companion, which was custom-made for translation to Cantonese by a community builder.\footnote{\url{https://chat.openai.com/share/7ee588af-dc48-4406-95f4-0471e1fb70a8}}

We used a range of different evaluation metrics including the lexical-based SacreBLEU \cite{post-2018-call} and hLEPOR \cite{han-etal-2013-language,han-etal-2021-cushlepor}, and the embedding-based BERTscore \cite{Zhang2020BERTScore} and COMET \cite{rei-etal-2020-comet}. 
hLEPOR has reported much higher correlation scores to the human evaluation than BLEU and other lexical-based metrics on the WMT shared task data \cite{han-etal-2013-description}. 
However, recent WMT metrics task findings have demonstrated the advantages of neural metrics based on embedding space similarities \cite{freitag-etal-2022-results}.




\begin{figure*}[h]
\centering
     \begin{subfigure}[]{0.3\textwidth}
         \centering
         \includegraphics[width=\textwidth]{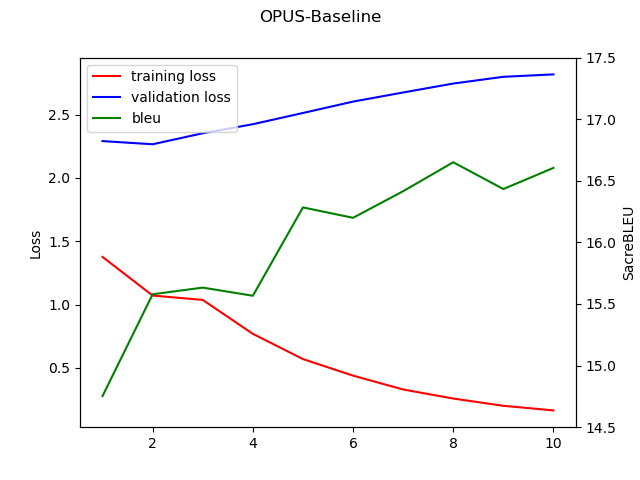}
         \caption{OPUS Baseline}
         \label{fig:opus-loss}
     \end{subfigure}
     \hfill
     \begin{subfigure}[]{0.3\textwidth}
         \centering
         \includegraphics[width=\textwidth]{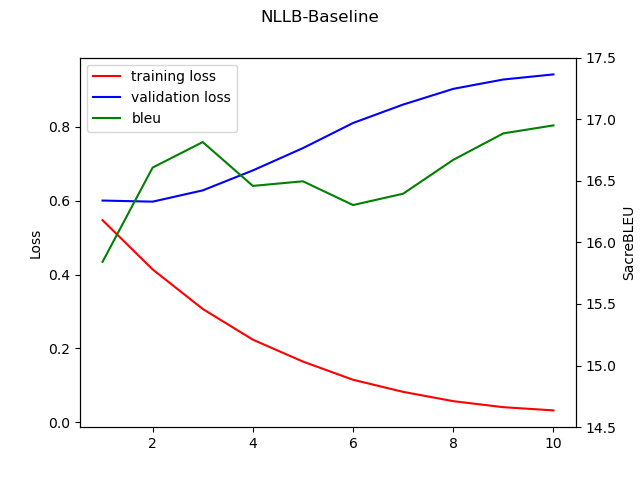}
         \caption{NLLB Baseline}
         \label{fig:nllb-loss}
     \end{subfigure}
     \hfill
     \begin{subfigure}[]{0.3\textwidth}
         \centering
         \includegraphics[width=\textwidth]{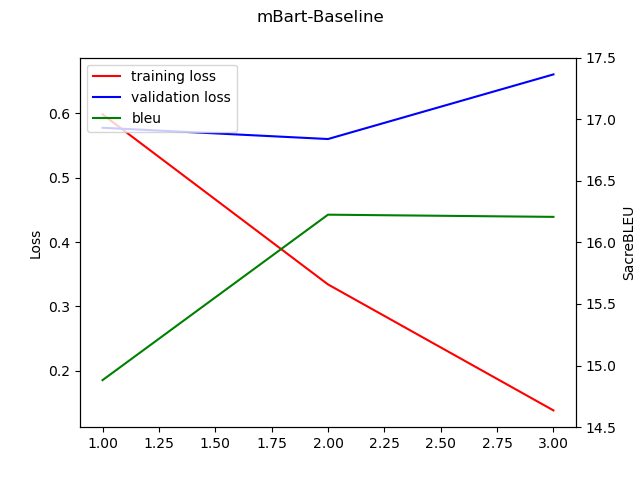}
         \caption{mBART Baseline}
         \label{fig:mbart-loss}
     \end{subfigure}
        \caption{Learning curves during model training using real data.}
    \label{fig:learning-curves}
\end{figure*}

\begin{table*}[!htb]
    \centering
    \begin{tabular}{l|cccc}
        \toprule
        Model Name & SacreBLEU & hLEPOR & BERTscore & COMET \\ \hline
        nllb-forward-bl & 16.5117 & 0.5651 & 0.9248 & 0.7376 \\
        nllb-forward-syn-h:h & 15.7751 & 0.5616 & 0.9235 & 0.7342 \\ 
        nllb-forward-syn-1:1 & \textbf{16.5901} & 0.5686 & \textbf{0.925} & \textbf{0.7409} \\ 
        nllb-forward-syn-1:1-10E & 16.5203 & \textbf{0.5689} & 0.9247 & 0.738 \\ 
        nllb-forward-syn-1:3 & 15.9175 & 0.5626 & 0.924 & 0.7376 \\ 
        nllb-forward-syn-1:5 & 15.8074 & 0.562 & 0.9237 & 0.7386 \\ \hline
        nllb-forward-syn-1:1-mbart & \textbf{16.8077} & \textbf{0.571} & \textbf{0.9256} & \textbf{0.7425} \\
        nllb-forward-syn-1:3-mbart & 15.8621 & 0.5617 & 0.9246 & 0.7384 \\
        nllb-forward-syn-1:1-opus & 16.5537 & 0.5704 & 0.9254 & 0.7416 \\
        nllb-forward-syn-1:3-opus & 15.9348 & 0.5651 & 0.9242 & 0.7374 \\ \hline
        mbart-forward-bl & 15.7513 & 0.5623 & 0.9227 & 0.7314 \\
        mbart-forward-syn-1:1-nllb & \textbf{16.0358} & \textbf{0.5681} & \textbf{0.9241} & \textbf{0.738} \\
        mbart-forward-syn-1:3-nllb & 15.326 & 0.5584 & 0.9225 & 0.7319 \\ \hline
        opus-forward-bl-10E & \textbf{15.0602} & \textbf{0.5581} & \textbf{0.9219} & \textbf{0.7193} \\
        opus-forward-syn-1:1-10E-nllb & 13.0623 & 0.5409 & 0.9164 & 0.6897 \\
        opus-forward-syn-1:3-10E-nllb & 13.3666 & 0.5442 & 0.9167 & 0.6957 \\ \hline
        baidu & 16.5669 & 0.5654 & 0.9243 & 0.7401 \\ 
        bing & 17.1098 & 0.5735 & 0.9258 & 0.7474 \\ 
        gpt4-ft(CantoneseCompanion) & \textbf{19.1622} & \textbf{0.5917} & \textbf{0.936} & \textbf{0.805} \\ \hline\hline 
        nllb-forward-bl-plus-wenlin14.5k & {\textit{16.6662}} & \underline{\textit{0.5828}} & \underline{\textit{0.926}} & \underline{0.7496} \\
        mbart-forward-bl-plus-wenlin14.5k & 15.2404 & 0.5734 & 0.9238 & 0.7411 \\
        opus-forward-bl-plus-wenlin14.5k & 13.0172 & 0.5473 & 0.9157 & 0.6882 \\ \hline\hline 
        nllb-200-deploy-no-finetune & 11.1827 & 0.4925 & 0.9129 & 0.6863 \\
        opus-deploy-no-finetune & 10.4035 & 0.4773 & 0.9082 & 0.6584 \\
        mbart-deploy-no-finetune & 8.3157 & 0.4387 & 0.9005 & 0.6273 \\ \hline\hline 
        nllb-forward-all3corpus & \underline{\textit{16.9986}} & \underline{\textit{0.583}} & \underline{\textit{0.927}} & \underline{\textit{0.7549}} \\
        nllb-forward-all3corpus-10E & 16.1749 & 0.5728 & 0.9254 & 0.7508 \\
        mbart-forward-all3corpus & 16.3204 & 0.5766 & 0.9253 & 0.7482 \\
        opus-forward-all3corpus-10E & 14.4699 & 0.5621 & 0.9191 & 0.7074 \\
        \bottomrule
    \end{tabular}
    \caption{Automatic Evaluation Scores from Different Models in\textsc{ CantonMT}. bl: bilingual real data; syn: synthetic data; h:h - half and half; 1:1/3/5 - 100\% real + 100/300/500\% synthetic; 10E: 10 epochs (default: 3); top-down second slot: model switch: model type using NLLB but synthetic data from other models (mBART and OpusMT); top-down third slot: including model switch for mBART fine-tuning using synthetic data generated from NLLB; similarly top-down forth slot: including model switch for OpusMT fine-tuning using synthetic data from NLLB. Bottom slot of Cluster 1: Bing/Baidu Translator and GPT4-finetuned Cantonese Companion; \textbf{bold} case is the best score of the same slot among the same model categories.
    Cluster 2: bilingual fine-tuned models using 38K words.hk data plus 14.5k Wenlin data; \textit{italic} indicates the number outperforms the same model fine-tuned with less data 38K.
    Cluster 3: Deployed Model without fine-tuning
    Cluster 4: Finetuned with the previous 2 corpora and an additional 10K data from OPUS Corpora we managed to find in the end - it shows the evaluation improvement continues.
    }
    \label{tab:Evaluation-metrics-scores}
\end{table*}

\subsection{Evaluations of \textsc{CantonMT}}

The learning curves of three base models during training using the 38K real data are shown in Figure \ref{fig:learning-curves} from left to right for mBART, NLLB-200 and Opus-MT. We used three epochs for mBART because it is too large for the computational resources available to us. From the learning curves, we can see that NLLB-200 has a peak score at epoch 3 then there is a dramatic drop until epoch 6, followed by an increase until epoch 10. In contrast, the Opus-MT model achieves a steady increase in its SacreBLEU score with more epochs, although there are little drops in between.

The automatic evaluation scores from \textsc{ CantonMT} models and other commercial engines are listed in Table \ref{tab:Evaluation-metrics-scores}.
Below are some interesting findings from the evaluation outcomes. 

\begin{itemize}
    \item LLM-ft vs -LLM-ft-syn: (1) NLLB-syn-1:1 has slightly better scores than NLLB-bl on all metrics, but increasing the ratio of synthetic data will decrease the scores such as in the 1:3 and 1:5 configurations, with around 1 absolute SacreBLEU point. (2) Similarly, mBART-syn-1:1 also outperforms mBART-ft but increasing the ratio of synthetic data will reduce the evaluation scores such as in the 1:3 configuration. (3) Surprisingly, the synthetic model for Opus-mt does not outperform Opus-ft-bl, which indicates that the quality of the generated synthetic data matters.
    \item Model Switching Matters: (1) the NLLB fine-tuned model using synthetic data from mBART (second model from the top of the table) produced higher scores than using the synthetic data generated from its own (first model from the top of the table). (2) mBART fine-tuned using NLLB-generated synthetic data also outperforms mBART fine-tuning using only bilingual real data. (3) In a similar situation, Opus-MT performs differently in comparison to the other two models.
    
        \item Commercial MT models: (1) GPT4-finetuned produced the highest evaluation scores but the free version of GPTs restricts the input number of strings; the data size used for fine-tuning GPT-4 is unknown and such data is not publicly available to researchers; 
        furthermore, it is unclear how GPT-4 performs MT; in addition, there are risks to data privacy when users choose to use engines from commercial companies. 
        In contrast, \textsc{CantonMT }is open-source, free, and researchers can continue to fine-tune it with their data or include more models, and is fully \textit{confidential} for users. 2) Bing and Baidu translators produced similar evaluation scores to the best system from \textsc{CantonMT}, though Bing produced slightly higher scores than Baidu, especially on the lexical-based metrics SacreBLEU and hLEPOR.

        \item Comparing to Model Deployment without Finetuning: in Cluster 3 (bottom) of Table~\ref{tab:Evaluation-metrics-scores}, model deployment without fine-tuning has much lower scores; these scores show that fine-tuning and synthetic data augmentation lead to a large increase in scores of around 50\% for all models using SecreBLEU.

\end{itemize}

\subsection{Adding More Real Data}
In the extension of our work, we managed to fine-tune the baseline models using more real data from another source called Wenlin\footnote{\url{https://wenlin.com/}} where we obtained another 14.5K parallel Cantonese-English dictionary.
We are curious about the model performance using more real data in addition to the 38K training corpus from words.hk. 
We listed the comparison scores in the second cluster of Table \ref{tab:Evaluation-metrics-scores} where it shows that the newly fine-tuned NLLB-200 using 52.5K data (38+14.5K=52.5K) produced higher scores on all metrics in comparison to 38K trained model; mBART fine-tuned using 52.5K obtains better scores on three metrics except for SacreBLEU; Opus-MT surprisingly did not get any increase across the metrics.
Nevertheless, these outcomes demonstrated the possibility of improving model performance with more available real data, at least for the NLLB and mBART models.
Moreover, \textbf{data quality matters}: simply adding 14.5K real data to fine-tune NLLB produced higher scores (\underline{underlined scores}) than the best synthetic system that used 38x2=76K data.
Subsequently, when we managed to get another 10K real data from Opus corpus, it shows continuous improvement by training using all three corpus we have, located in the Cluster 4 bottom of the table.

\begin{figure*}[t]
\centering
\includegraphics[width=0.75\textwidth]{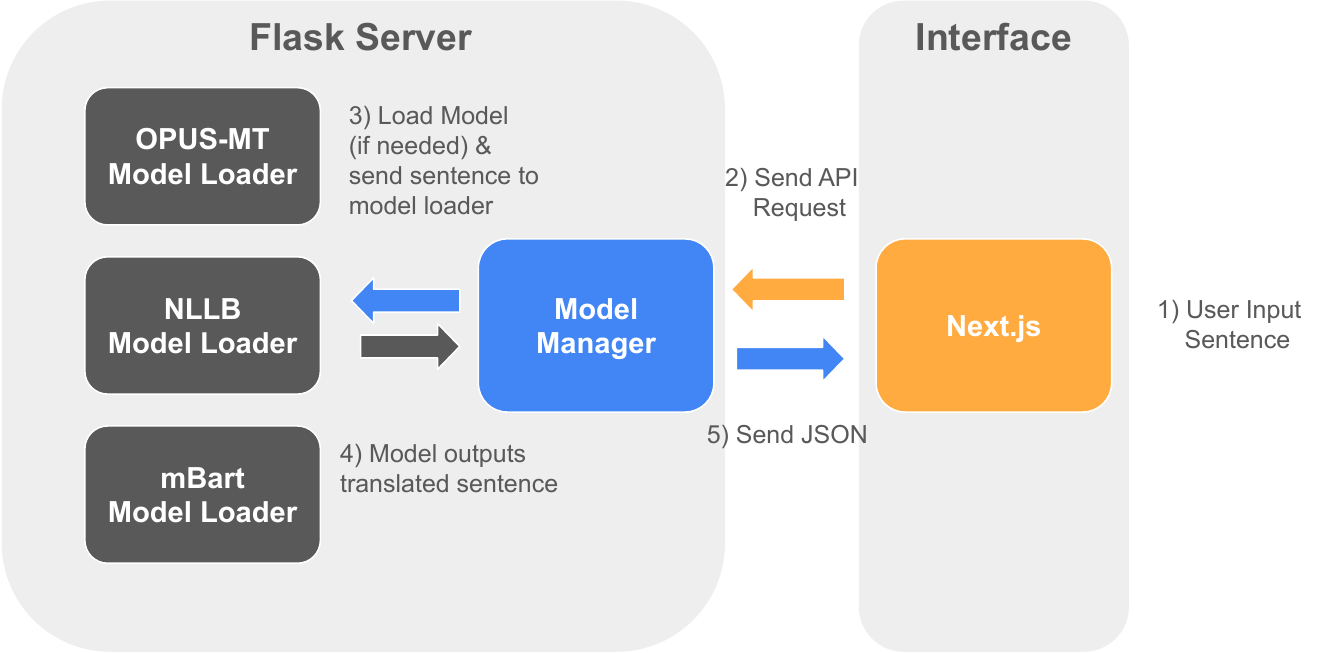}
    \caption{\textsc{CantonMT} Server and Interface Flowchart diagram.}
    \label{fig:User-Interface-Flowchart-cropped}
\end{figure*}


\section{\textsc{CantonMT }Platform}
\label{sec:cantonMT}
To further facilitate Cantonese-English MT research and for users to easily access freely available fine-tuned models, we developed a user-friendly interface for the CantonMT platform. 
Users can choose different models and translation directions (Cantonese$\Leftrightarrow$English) via the interface (Figure \ref{fig:screenshot_CantonMT} in the Appendix). 
The 
web application contains two main parts, the Interface and the Server.

\subsection{User Interface}
To test the user interface and different models for translation, users can choose from different model types and source languages, which dynamically capture the available models in the server, and allow users to select different training methods for the model. One can then type the source sentence in the input box and click the ``Translate'' button to obtain the translation output from the model.
The application layout is quite modular in case different model types or languages are added to the system, which could potentially be used as a base framework for different translation systems. It is possible to simply add more languages to the input and output if one wishes to expand the implementations.
%
The look-and-feel of this web application is based on a template \citep{ai-code-translator} for an AI Code translator, which was customised and developed in TypeScript with the Next.js framework. The reason for choosing this framework is that it provides a very modern and minimalistic approach. 

\subsection{Server}
A diagram outlining the modules can be seen in Figure \ref{fig:User-Interface-Flowchart-cropped} to understand the general structure of the server.
Users can easily run the server on their local machines by following the instructions provided in a README file.
The server has two main functionalities, where the first one will output the list of model paths given the model type and source languages. The second one provides the translation, where one could provide the details of the model and also the sentence in the language specified, and the server would respond with the translated sentence using the model output.

During our implementation, due to memory constraints, the server crashed multiple times on our local machine. To mitigate the risk of server crashes, a \textit{model manager} was produced, which implements a Least Recently Used (LRU) cache for the different model loaders, where the least recently used model will be deleted from memory if it exceeds the limit of the number of models.
%
%
The server is built entirely based upon the \textit{Python Flask} library. The reason for choosing this framework is that the models can be run on top of the Python Transformers library, which provides seamless implementation without much additional effort.

\section{Discussion and Conclusion}
\label{sec:discuss}

In this work, we investigated the back-translation methodology for bilingual synthetic data generation for the sake of data augmentation for NMT, on a new language translation direction, Cantonese-to-English. We tested both smaller-sized OpusMT and extra-large LLMs NLLB and mBART both using available bilingual real data and larger synthetic data.
Our experiments show that all the fine-tuned models outperformed the baseline deployment models with large margins. Furthermore, the synthetic model nllb-syn-1:1-mbart produced higher scores using the model switch method compared to those without the model switch. 
Lastly, the best performing fine-tuned models have similar (or even higher) evaluation scores than the current commercially available translators of Baidu and Microsoft-Bing. 

In terms of concerns of \textbf{data privacy} such as handling of sensitive data (e.g., in clinical applications related to health analytics of patient data \cite{han6neural_Frontiers}), \textsc{CantonMT} can be fully controlled by users without interference from any third parties.
We open-source our platform so that researchers can continue to integrate new models into the toolkit to promote Cantonese-English MT.
We also plan to carry out human evaluations on the outputs from different systems to get more insights into the system errors.


\section*{Limitations}
The synthetic data generated in this work is based on the fine-tuned model using 38K words.hk bilingual dictionary corpus, the first corpus we managed to find. This restricted the synthetic data quality. In the following-up work, we plan to use the further fine-tuned model on all three corpora, words.hk, wenlin, and opus-10K, to generate better back-translated synthetic data. We expect this will improve the synthetic data fine-tuned models.

The whole procedure of how difficult it was to collect real Cantonese-English bilingual data shows that Cantonese-English MT is still at its beginning stage with many obstacles and challenges to public research. 

\section*{Acknowledgements}

We thank words.hk and wenlin.com for the data.
LH and GN thank the support from the following
grants: (1) ``Assembling the Data Jigsaw: Powering Robust Research on the Causes, Determinants and Outcomes of MSK Disease''. This project has been funded by the Nuffield Foundation (www.nuffieldfoundation.org), but the views expressed are those of the authors and not necessarily of the Foundation; and  
(2) the UKRI/EPSRC Grant EP/V047949/1 ``Integrating hospital outpatient letters into the healthcare data space''.

\begin{figure*}[t]
\centering
\includegraphics[width=0.99\textwidth]{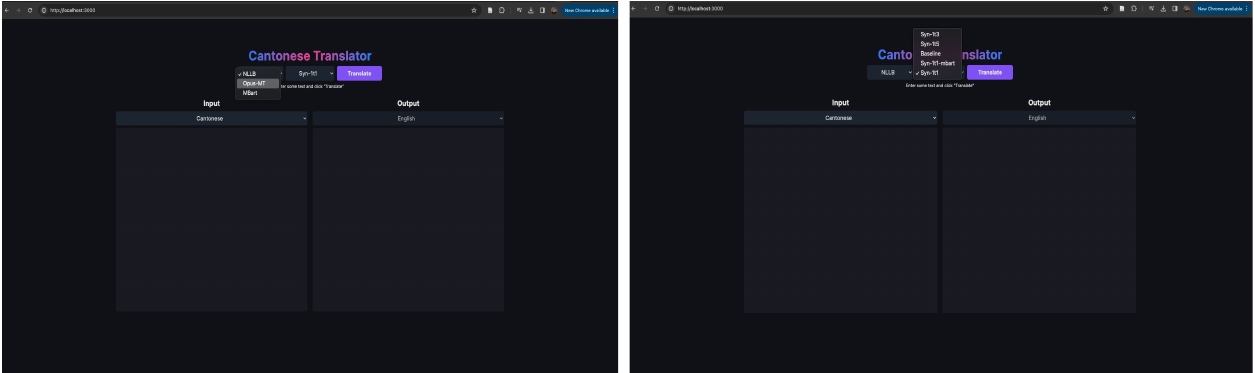}
    \caption{\textsc{CantonMT} Platform with options of model types, training categories, and translating directions.  Frontend:  TypeScript with Next.js. Backend: Python - Flask  }
    \label{fig:screenshot_CantonMT}
\end{figure*}

\begin{figure*}[t]
\centering
\includegraphics[width=0.99\textwidth]{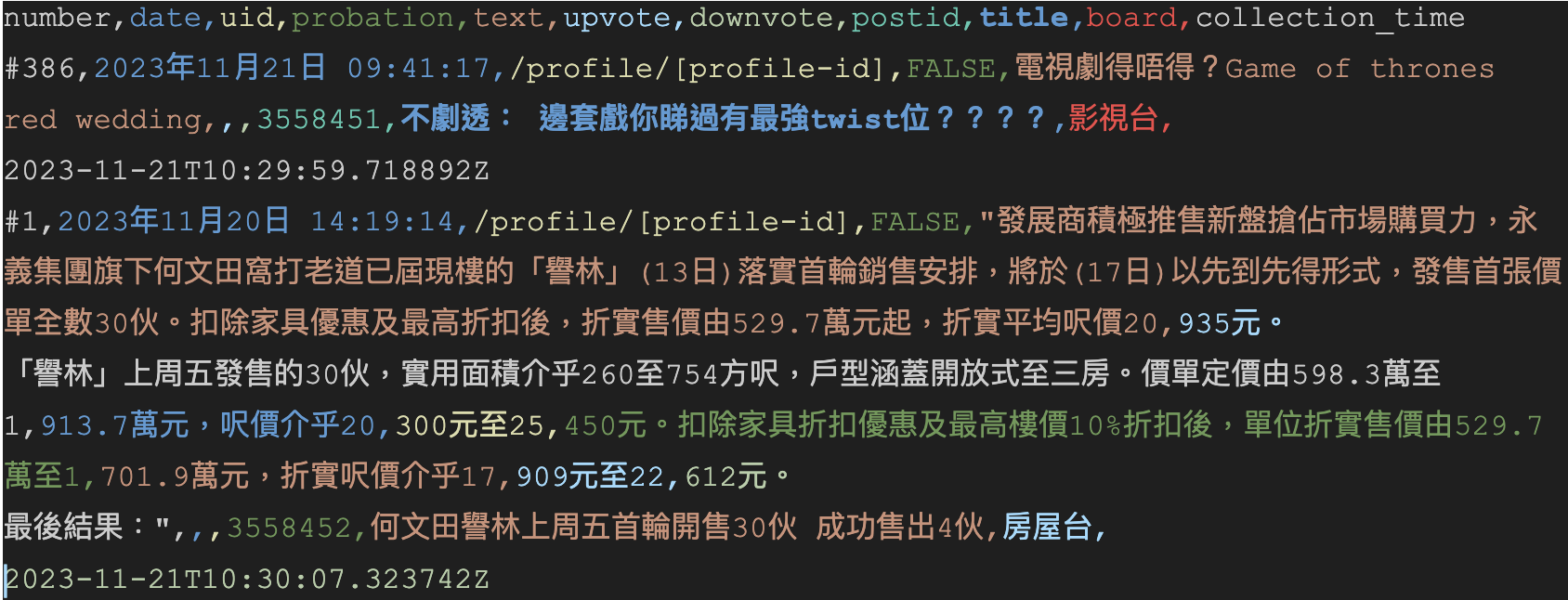}
    \caption{Example text extracted from LIHKG website with lots noise before cleaning and anonymisation}
    \label{fig:LIHKG-example}
\end{figure*}

\bibliography{anthology,custom}

\appendix

\section{ Appendix}
\label{sec:appendix}
Example raw text extracted from LIHKG website can be seen in Figure \ref{fig:LIHKG-example} before cleaning.
CantonMT user-friendly interface is shown in Figure \ref{fig:screenshot_CantonMT} (Frontend:  TypeScript with Next.js. Backend: Python - Flask).

The model parameters from OpusMT, extra-large NLLB and mBART are shown in Table \ref{fig:models-parameter-set-cropped}, which shows that NLLB and mBART have doubled the number of transformer layers and have almost 10 times more parameters than OpusMT.
\begin{table}[!ht]
    \centering
    \begin{tabular}{llll}
        \toprule
        ~ & Opus & NLLB & mBart \\ \hline
        Layers & 12 & 24 & 24 \\
        Hidden Unit & 512 & 1024 & 1024 \\
        Model Parameters & 77.9M & 615M & 610.9M \\
        Language Pair & No & Yes & No \\
        Release Year & 2020 & 2022 & 2020 \\ \bottomrule
    \end{tabular}
    \caption{Parameters from deployed models.
    Language pair: if the model contains Cantonese-English as a language pair}
    \label{fig:models-parameter-set-cropped}
\end{table}

Explanation of Abbriviations used in the scoring table:
\begin{itemize}
    \item ``nllb-forward-bl'': NLLB fine-tuned model in the forward translation direction (Cantonese-English) using the real 38K bilingual corpus 
    \item ``nllb-forward-syn-h:h'': NLLB fine-tuned model using forward-translation generated synthetic data to substitute half of the 38K real data, i.e. 19K real and 19K synthetic 
    \item ``nllb-forward-syn-1:1'': NLLB fine-tuned model using forward-translation generated synthetic data with the ratio 1:1, i.e. 38K real and 38K synthetic
    \item ``nllb-forward-syn-1:1-10E'': the same with above corpus setting but running 10 epochs, default is 3 epochs only
    \item ``nllb-forward-syn-1:1-mbart'': NLLB model fine-tuning using forward-translation generated synthetic data by another model mBART, 38K real and 38K synthetic
\end{itemize}

\begin{table*}[!htb]
    \centering
    \begin{tabular}{l|cccc}
        \toprule
        Model Name & SacreBLEU & BERTscore & COMET \\ \hline
        mBART-back-bl & \textbf{20.3841} & \textbf{0.7944} & \textbf{0.8095} \\
        mBART-back-syn-1:1-NLLB+ & 20.1923 & 0.7921 & 0.8068 \\ \hline
        nllb-back-bl & 18.4713 & 0.7877 & 0.7927 \\
        nllb-back-syn-1:1 & 17.9400 & 0.7807 & 0.7772 \\
        nllb-back-syn-1:3 & 12.0352 & 0.7628 & 0.7493 \\ \hline
        opus-back-bl & 18.1496 & 0.7811 & 0.7816 \\
        opus-back-syn-1:1-NLLB+ & 17.9346 & 0.7781 & 0.7715 \\
        \bottomrule
    \end{tabular}
    \caption{Automatic Evaluation Scores from Different Models in\textsc{ CantonMT}.}
    \label{tab:Evaluation-metrics-scores}
\end{table*}

\begin{table}[!htb]
    \centering
    \begin{tabular}{|l|l|}
        \hline
        Hyperparameter & Value \\
        \hline
        Learning Rate & $10_{-4}$ \\
        Weight Decay & 0.01 \\
        FP16 & True \\
        \hline
    \end{tabular}
    \caption{Fine-tuning Hyperparameters w/ Hugging Face Trainer API}
    \label{tab:finetune_hyperparameters}
\end{table}

\end{document}